\def\year{2021}\relax
\documentclass[letterpaper]{article} % DO NOT CHANGE THIS
\usepackage{aaai21}  % DO NOT CHANGE THIS
\usepackage{times}  % DO NOT CHANGE THIS
\usepackage{helvet} % DO NOT CHANGE THIS
\usepackage{courier}  % DO NOT CHANGE THIS
\usepackage[hyphens]{url}  % DO NOT CHANGE THIS
\usepackage{graphicx} % DO NOT CHANGE THIS
\usepackage{natbib}  % DO NOT CHANGE THIS AND DO NOT ADD ANY OPTIONS TO IT
\usepackage{caption} % DO NOT CHANGE THIS AND DO NOT ADD ANY OPTIONS TO IT
\usepackage[utf8]{inputenc} % allow utf-8 input
\usepackage[T1]{fontenc}    % use 8-bit T1 fonts
\usepackage{url}            % simple URL typesetting
\usepackage{booktabs}       % professional-quality tables
\usepackage{amsfonts}       % blackboard math symbols
\usepackage{nicefrac}       % compact symbols for 1/2, etc.
\usepackage{microtype}      % microtypography
\usepackage{amsmath}
\usepackage[linesnumbered,ruled,vlined]{algorithm2e}
\usepackage{multirow}
\usepackage{multicol}
\usepackage{subfigure}
\usepackage{color}
\newcommand{\domX}{\mathcal{X}}
\newcommand{\domY}{\mathcal{Y}}
\newcommand{\Ddev}{D_{\text{valid}}}
\newcommand{\Dtrain}{D_{\text{train}}}

\DeclareMathOperator*{\argmin}{\arg\!\min}
\usepackage[switch]{lineno}  %
\title{Learning to Reweight with Deep Interactions\thanks{This work was done when Yang Fan was an intern at Microsoft Research Asia. }}
\author{
    Yang Fan\textsuperscript{\rm 1},
    Yingce Xia\textsuperscript{\rm 2},
    Lijun Wu\textsuperscript{\rm 2}, 
    Shufang Xie\textsuperscript{\rm 2}, \\
    Weiqing Liu\textsuperscript{\rm 2},
    Jiang Bian\textsuperscript{\rm 2},
    Tao Qin\textsuperscript{\rm 2},
    Xiang-Yang Li\textsuperscript{\rm 1} \\
}
\affiliations{
    \textsuperscript{\rm 1}University of Science and Technology of China\quad
    \textsuperscript{\rm 2}Microsoft Research Asia\\
    fyabc@mail.ustc.edu.cn, xiangyangli@ustc.edu.cn\\
    \{yingce.xia, lijuwu, shufxi, Weiqing.Liu, Jiang.Bian, taoqin\}@microsoft.com,
    
}

\begin{document}
%\linenumbers
\maketitle

\begin{abstract}
Recently, the concept of teaching has been introduced into machine learning, in which a teacher model is used to guide the training of a student model (which will be used in real tasks) through  data selection, loss function design, etc. Learning to reweight, which is a specific kind of teaching that reweights training data using a teacher model, receives much attention due to its simplicity and effectiveness. In existing learning to reweight works, the teacher model only utilizes shallow/surface information such as training iteration number and loss/accuracy of the student model from training/validation sets, but ignores the internal states of the student model, which limits the potential of learning to reweight. In this work, we propose an improved data reweighting algorithm, in which the student model provides its internal states to the teacher model, and the teacher model returns adaptive weights of training samples to enhance the training of the student model. The teacher model is jointly trained with the student model using meta gradients propagated from a validation set. Experiments on image classification with clean/noisy labels and neural machine translation empirically demonstrate that our algorithm makes significant improvement over previous methods.
\end{abstract}

\section{Introduction}
% Inspired by human education systems, the concept of teaching has been introduced into machine learning, in which a teacher model is employed to teach and assist the training of a student model. Previous work can be categorized into two branches: (1) The teacher and student share the same input and output spaces~\cite{zhu2015machine,zhu2016teachingdim} and the teacher model aims to teach the student model with minimal cost, like data selection~\cite{liu2017iterative,liu2018towards} in supervised learning and action modification in reinforcement learning~\cite{zhang2020teaching}. (2) The teacher and student models are of different input/output spaces~\cite{fan2018dataTeach,wu2018lossTeach}: The student model is used for real tasks (e.g., image classification, machine translation); the teacher model takes the information from the student model and the validation set as inputs and outputs some signals to guide the training of the student model, e.g., adjusting the weights of training data~\cite{fan2018dataTeach,metaweightnet,jiang2018mentornet,ren2018learning}, generating better loss functions~\cite{wu2018lossTeach}, etc.
Inspired by human education systems, the concept of teaching has been introduced into machine learning, in which a teacher model is employed to teach and assist the training of a student model. Previous work can be categorized into two branches: (1) One is to transfer knowledge (e.g., image classification, machine translation) from the teacher to student~\cite{zhu2015machine,zhu2016teachingdim} and the teacher model aims to teach the student model with minimal cost, like data selection~\cite{liu2017iterative,liu2018towards} in supervised learning and action modification in reinforcement learning~\cite{zhang2020teaching}. (2) The student model is used for real tasks (e.g., image classification, machine translation) and the teacher is a meta-model that can guide the training the of student model. The teacher model takes the information from the student model and the validation set as inputs and outputs some signals to guide the training of the student model, e.g., adjusting the weights of training data~\cite{fan2018dataTeach,metaweightnet,jiang2018mentornet,ren2018learning}, generating better loss functions~\cite{wu2018lossTeach}, etc.
%Another branch, which is known as learning to teach~\cite{fan2018dataTeach,wu2018lossTeach}, is a meta-learning paradigm where the teacher guides the training the of student model. The student model is used for real tasks (e.g., image classification, machine translation), while the teacher model cannot handle those tasks. The teacher model integrates the information from the student model and the validation signal and then updates the training procedure, like adjusting the weights of training data~\cite{fan2018dataTeach,metaweightnet,jiang2018mentornet,ren2018learning}, generating better loss functions~\cite{wu2018lossTeach}, etc.
%involves a teacher model and a student model. The student model is our final target and used for real tasks like image classification and machine translation, while the teacher model is used to guide the training of the student model through training data selection~\cite{liu2017iterative}, adjusting the weights of training data~\cite{fan2018dataTeach,metaweightnet,jiang2018mentornet,ren2018learning}, generating better loss functions~\cite{wu2018lossTeach}, etc. 
These approaches have shown promising results in image classification~\cite{jiang2018mentornet,metaweightnet}, machine translation~\cite{wu2018lossTeach}, and text classification~\cite{fan2018dataTeach}. Among the teaching methods, learning to reweight the data is widely adopted due to its simplicity and effectiveness and we focus on this direction in this work. 

Previously, the teacher model used for data reweighting only utilizes surface information derived from the student model. In~\cite{fan2018dataTeach,wu2018lossTeach,metaweightnet,jiang2018mentornet}, the inputs of the teacher model include training iteration number, training loss (as well as the margin~\cite{schapire1998boosting}), validation loss, the output of the student model, etc. In those algorithms, the teacher model does not leverage the internal states of the student model, e.g., the values of the hidden neurons of a neural network based student model. We notice that the internal states of a model have been widely investigated and shown its effectiveness in deep learning algorithms and tasks. In ELMo~\cite{peters-etal-2018-deep}, a pre-trained LSTM provides its internal states, which are the values of its hidden layers, for downstream tasks as feature representations. In image captioning~\cite{xu2015show,anderson2018bottom}, a faster RCNN~\cite{NIPS2015_5638} pre-trained on ImageNet provides its internal states (i.e., mean-pooled convolutional features) of the selected regions, serving as representations of images~\cite{anderson2018bottom}. In knowledge distillation~\cite{romero2015fitnets,aguilar2019knowledge}, a student model mimics the output of the internal layers of the teacher model so as to achieve comparable performances with the teacher model. However, to the best of our knowledge, this kind of deep information is not extensively investigated in learning to reweight algorithms. The success of leveraging internal states in the above algorithms and applications motivates us to investigate them in learning to reweight, which leads to deep interactions between the teacher and student model.

\begin{figure}[!htbp]
\centering
\includegraphics[width=0.745\linewidth]{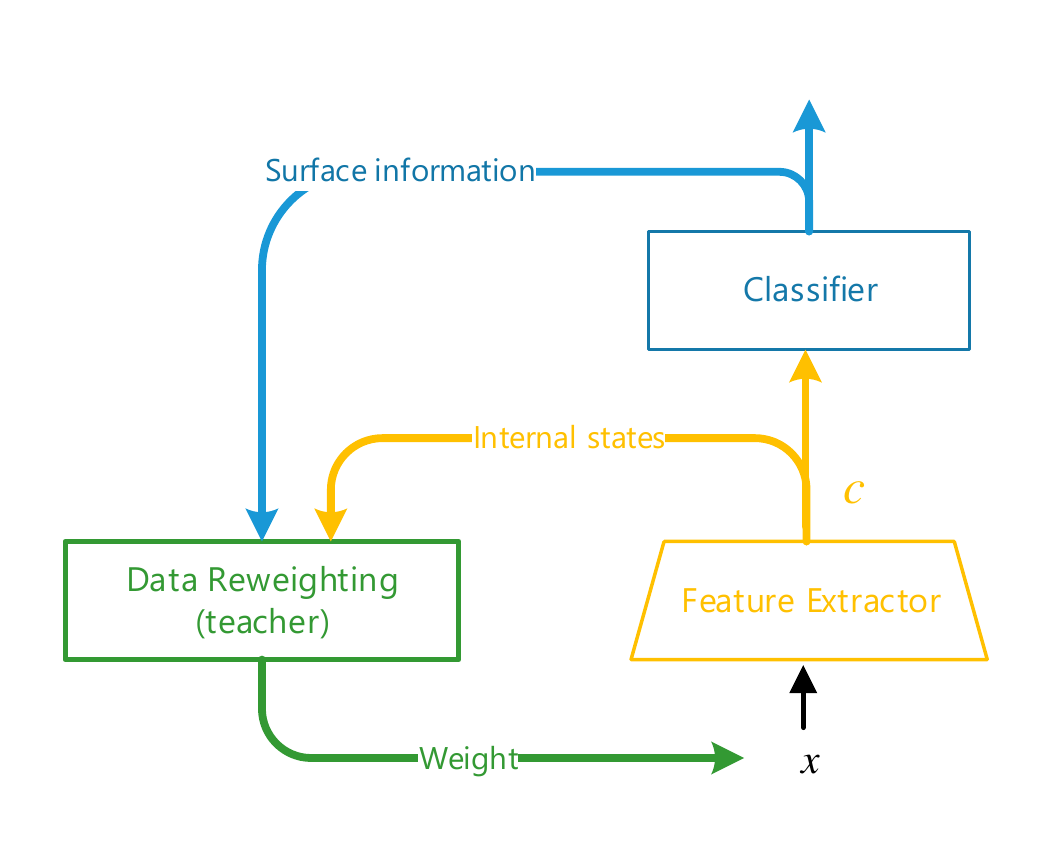}
\caption{Workflow of our approach.}
\label{fig:net_comp}
\end{figure}

We propose a new data reweighting algorithm, in which the teacher model and the student model have deep interactions: the student model provides its internal states (e.g., the values of its internal layers) and optionally surface information (e.g., predictions, classification loss) to the teacher model, and the teacher model outputs adaptive weights of training samples which are used to enhance the training of the student model. A workflow of our method is in Figure~\ref{fig:net_comp}. 
%illustrates the key difference between our algorithm (the right figure) and previous data reweighting algorithm~\cite{fan2018dataTeach,wu2018lossTeach}  (the left figure). 
We decompose the student model into a feature extractor, which can process the input $x$ to an internal state $c$ (the yellow parts), and a classifier, which is a relatively shallow model (e.g., a linear classifier; blue parts) to map $c$ to the final prediction $\hat{y}$. In~\cite{fan2018dataTeach,wu2018lossTeach}, the teacher model only takes the surface information of the student model as inputs like training and validation loss (i.e., the blue parts), which are related to $\hat{y}$ and ground truth label $y$ but not explicitly related to the internal states $c$.
% \footnote{While the teacher model also has some other inputs such as training iteration number and validation loss, they are not directly related to the internal states of the student model.}, 
In contrast, the teacher model in our algorithm leverages both the surface information  and the internal states $c$ of the student model as inputs. In this way, more information from the student model becomes accessible to the teacher model. 

In our algorithm, the teacher and the student models are jointly optimized in an alternating way, where the teacher model is updated according to the validation loss via reverse-mode differentiation~\cite{maclaurin2015gradient}, and the student model tries to minimize the loss on reweighted data. Experimental results on CIFAR-$10$ and CIFAR-$100$~\cite{krizhevsky2009learning} with both clean labels and noisy labels demonstrate the effectiveness of our algorithm. We also conduct a group of experiment on IWSLT German$\to$Englsih neural machine translation to demonstrate the effectiveness of our method on sequence generation tasks. We achieve promising results over previous methods of learning to teach. 

% Specifically, we achieve state-of-the-art results on CIFAR-$10$/$100$ classification with noisy labels.

\section{Related Work}\label{sec:related_work}
Assigning weights to different data points have been widely investigated in literature, where the weights can be either continuous~\cite{friedman2000additive,jiang2018mentornet} or binary~\cite{fan2018dataTeach,bengio2009curriculum}.  The weights can be explicitly bundled with data, like Boosting and AdaBoost methods~\cite{freung1997decision,hastie2009multi,friedman2000additive} where the weights of incorrectly classified data are gradually increased, or implicitly achieved by controlling the sampling probability, like hard negative mining~\cite{malisiewicz2011ensemble} where the harder examples in a previous round will be sampled again in the next round. As a comparison, in self-paced learning (SPL)~\cite{Kumar2010SPL}, weights of hard examples will be assigned to zero in the early stage of training, and the threshold is gradually increased during the training process to control the student model to learn from easy to hard. An important motivation of data weighting is to increase the robustness of training, including addressing the problem of imbalanced data~\cite{SUN20073358,dong2017class,8012579}, biased data~\cite{zadrozny2004learning,ren2018learning}, noisy data~\cite{angluin1988learning,reed2014training,sukhbaatar2014learning,koh2017understanding}. The idea of adjusting weights for the data is also essential in another line of research about combing optimizers with different sampling techniques~\cite{katharopoulos2018not, liu2020adam, namkoong2017adaptive}.

Except for manually designing weights for the data, there is another branch of work that leverages a meta model to assign weights. Learning to teach~\cite{fan2018dataTeach} is a learning paradigm where there is a student model for the real task, and a teacher model to guide the training of the student model. Based on the collected information, the teacher model provides signals to the student model, which can be the weights of training data~\cite{fan2018dataTeach}, adaptive loss functions~\cite{wu2018lossTeach}, etc. The general scheme of machine teaching is discussed and summarized in \cite{zhu2015machine}. The concept of teaching can be found in label propagation~\cite{gong2016label,gong2016teaching}, pedagogical teaching~\cite{ho2016showing,shafto2014rational}, etc. \cite{liu2017iterative} leverages a teaching way to speed up the training, where the teacher model selects the training data balancing the trade off between the difficulty and usefulness of the data. \cite{metaweightnet,ren2018learning,jiang2018mentornet} mainly focuses on the setting that the data is biased or imbalanced. %Generally, these methods are about using sampling methods to improve the models. The internal states of the models and the feedback signals from the validation set are not fully leveraged. Our method and these methods are complementary to each other.}

In machine teaching literature (which focuses on transferring knowledge from teacher models to student models), there are some works that that teacher get more information beyond surface information like loss function. \cite{liu2017iterative,liu2018towards} focuses on data selection to speed up the learning of student model. However, their algorithms and analysis are built upon linear models. \cite{lessard2019optimal} tackles a similar problem, which is about to find the shortest training sequence to drive the student model to the target one. According to our knowledge, there is no extensive study that teacher and student can deeply interact 
for data reweighting based on deep neural neural networks. We will empirically verify the benefits of our proposals.

\section{Our Method}\label{sec:our_method}
We focus on data teaching in this work, where the teacher model assigns an adaptive weight to each sample. We first introduce the notations used in this work, then describe our algorithm, and finally provide some discussions.

\subsection{Notations}
Let $\domX$ and $\domY$ denote the source domain and the target domain respectively. We want to learn a mapping $f$, i.e., the student model, from $\domX$ and $\domY$. W.l.o.g, we can decompose $f$ into a feature extractor and a decision maker, denoted as $\varphi_f$ and $\varphi_d$ respectively, where $\varphi_f:\domX\mapsto\mathbb{R}^d$, $\varphi_d:\mathbb{R}^d\mapsto\domY$, and $d$ is the dimension of the extracted feature. That is, for any $x\in\domX$, $f(x)=\varphi_d(\varphi_f(x))$. We denote the parameters of $f$ as $\theta$. %In our work, we mainly work on the image classification problem. 
In this section, we take image classification as an example and present our algorithm. Our approach can be easily adopted into other tasks like sequence generation, which is shown at Section~\ref{sec:exp_cifar_noise}. 

Given a classification network $f$, we manually define $\varphi_f(\cdot)$ is the output of the second-to-last layer, and $\varphi_d$ is a linear classifier taking $\varphi_f(x)$ as input. 
% the default segmentation {\color{red} Reviewer: segmentation?} method is that  is the output of the second-to-last layer, and $\varphi_d$ is a linear classifier taking $\varphi_f(x)$ as input. 
Let $\phi(I,M;\omega)$ denote the teacher model parameterized by $\omega$, where $I$ is the internal states of a student model and $M$ is the surface information like training iteration, training loss, labels of the samples, etc. 
%In the remaining text, we will replace $I$ with $\varphi_f(x)$ for any $x\in\domX$. 
$\phi$ can map an input sample $(x,y)\in\domX\times\domY$ to a non-negative scalar, representing the weight of the sample. Let $\ell(f(x),y;\theta)$ denote the training loss on sample pair $(x,y)$, and $R(\theta)$ is a regularization term on $\theta$, independent of the training samples. 

Let $\Dtrain$ and $\Ddev$ denote the training and validation sets respectively, both of which are subsets of $\domX\times\domY$ with $N_T$ and $N_V$ samples. Denote the validation metric as $m(y,\hat{y})$, where $y$ and $\hat{y}$ are the ground truth label and predicted label respectively. We require that $m(\cdot,\cdot)$ should be a differentiable function w.r.t. the second input. $m(y,\hat{y})$ can be specialized as the expected accuracy~\cite{wu2018lossTeach} or the log-likelihood on the validation set. Define $\mathcal{M}(\Ddev;\theta)$ as $\frac{1}{N_V}\sum_{(x,y)\in\Ddev}m(y,f(x;\theta))$.

\subsection{Algorithm}
The teacher model outputs a weight for any input data. When facing a real-world machine learning problem, we need to fit a student model on the training data, select the best model according to validation performance, and apply it to the test set. Since the test set is not accessible during training and model selection, we need to maximize the validation performance of the student model. This can be formulated as a bi-level optimization problem: 
%$\max_{\omega}M(\Ddev;\theta^*)$, where $\theta^*$ is obtained by 
%\begin{equation}
\begin{align}
&\max_{\omega,\theta^*(\omega)}\mathcal{M}(\Ddev;\theta^*(\omega))\nonumber\\
\text{s.t. } & \theta^*(\omega)=\argmin_\theta \frac{1}{N_T}\sum_{i=1}^{N_T}w(x_i,y_i)\ell(f(x_i),y_i;\theta)+\lambda R(\theta),\nonumber\\
&w(x_i,y_i)=\phi(\varphi_f(x_i),M;\omega),\label{eq:high_level_obj}
\end{align}
%\end{equation}
where $\lambda$ is a hyperparameter, and $w(x_i)$ represents the weight of data $x_i$. The task of the student model is to minimize the loss on weighted data, as shown in the second line of Eqn.\eqref{eq:high_level_obj}. Without a teacher, all $w(x_\cdot)$'s are fixed as one. In a learning-to-teach framework, the parameters of the teacher model (i.e., $\omega$) and the student model (i.e., $\theta$) are jointly optimized. Eqn.\eqref{eq:high_level_obj} is optimized in an iterative way, where we calculate $\theta^*(\omega)$ based on a given $\omega$, then we update $\omega$ based on the obtained $\theta^*(\omega)$. We need to figure out how to obtain $\theta^*$, and how to calculate $\frac{\partial }{\partial \omega}\mathcal{M}(\Ddev;\theta^*(\omega))$. 

%In previous work~\cite{metaweightnet,fan2018dataTeach,jiang2018mentornet}, since $w(x_i)$ is calculate with meta information only like different type of losses, they set $\frac{\partial w(x_i)}{\partial \theta}$ as zero. In our work, $w(x_i)$ explicitly relates to a representation provided by the student network, therefore when updating the student model, we should also leverage the gradient passed through $w(x_i)$ to update the student model. 

\noindent{\bf Obtaining $\theta^*(\omega)$}: Considering a deep neural network is highly non-convex, generally, we are not able to get the closed-form solution of the $\theta^*$ in Eqn.\eqref{eq:high_level_obj}. We choose stochastic gradient descend method (briefly, SGD) with momentum for optimization~\cite{polyak1964some}, which is an iterative algorithm. We use a subscript $t$ to denote the $t$-th step in optimization. $D_t$ is the data of the $t$-th minibatch, with the $k$-th sample $(x_{t,k},y_{t,k})$ in it. For ease of reference, denote $w_t$ as a column vector, where the $k$-th element is the weight for sample $(x_{t,k},y_{t,k})$, and $\ell(D_t;\theta_t)$ is another column vector with the $k$-element $\ell(f(x_{t,k}),y_{t,k};\theta_t)$, both of which are defined in Eqn.\eqref{eq:high_level_obj}. Following the implementation of PyTorch~\cite{NEURIPS2019_9015}, the update rule of momentum SGD is:
\begin{equation}
\begin{aligned}
v_{t+1} &= \mu v_t + \frac{\partial }{\partial \theta_t}\big[\frac{1}{\vert D_t\vert}w_t^\top \ell\big(D_t;\theta_t)+\lambda R(\theta_t)\big];\\
\theta_{t+1}&=\theta_t - \eta_t v_{t+1},
\end{aligned}
\label{eq:msgd_update_rule}
\end{equation}
where $\theta_0=v_0=0$. $\eta_t$ is the learning rate at the $t$-th step, and $\mu$ is the momentum coefficient. Assume we update the model for $K$ steps. We can eventually obtain $\theta_K$, which serves as the proxy for $\theta^*$. To stabilize the training, we will set $\frac{\partial w_t}{\partial\theta_t}=0$.

\noindent{\bf Calculating }$\frac{\partial }{\partial \omega}\mathcal{M}(\Ddev;\theta_K)$: Motivated by reverse-mode differentiation~\cite{maclaurin2015gradient}, we use a recursive way to calculate gradients. For ease of reference, let $\mathrm{d}\theta_t$ and $\mathrm{d}v_t$ denote  $\frac{\partial }{\partial \theta_t}\mathcal{M}(\Ddev;\theta_K)$ and $\frac{\partial }{\partial v_t}\mathcal{M}(\Ddev;\theta_K)$ respectively. According to the chain rule to compute derivative, for any $t\in\{0,1,2,\cdots,K-1\}$, we have
\begin{align}
\mathrm{d}\theta_t &= \big(\frac{\partial \theta_{t+1}}{\partial \theta_t}\big)^\top\mathrm{d}\theta_{t+1} + \big(\frac{\partial v_{t+1}}{\partial \theta_t}\big)^\top\mathrm{d}v_{t+1}\\
&=\mathrm{d}\theta_{t+1} + \frac{\partial^2}{\partial \theta_t^2}\big[\frac{1}{\vert D_t\vert}w_t^\top \ell\big(D_t;\theta_t)+\lambda R(\theta_t)\big]\mathrm{d}v_{t+1};\nonumber\\
\mathrm{d}v_t &= \big(\frac{\partial \theta_{t}}{\partial v_t}\big)^\top\mathrm{d}\theta_{t} + \big(\frac{\partial v_{t+1}}{\partial v_t}\big)^\top\mathrm{d}v_{t+1}\\
&=-\eta_t\mathrm{d}\theta_{t} + \mu\mathrm{d}v_{t+1};\label{eq:derivation}\\
\frac{\partial v_{t+1}}{\partial \omega}&=\frac{\partial^2}{\partial\theta_t\partial\omega}w_t^\top\ell(D_t;\theta_t);\\ &\frac{\partial}{\partial \omega}\mathcal{M}(\Ddev;\theta_K)=\sum_{t=1}^{K}(\frac{\partial v_t}{\partial \omega})^\top\mathrm{d}v_t.\nonumber
%\frac{\partial v_{t+1}}{\partial \omega}&=\frac{\partial^2}{\partial\theta_t\partial\omega}w_t^\top\ell(D_t;\theta_t).\nonumber
\end{align}
According to Eqn.\eqref{eq:derivation}, we can design Algorithm~\ref{alg:teacher_gradient} to calculate the gradients of the teacher model.

\begin{algorithm}[!h]
{\em Input}: Teacher model backpropagation interval $B$; parameters and momentum of the student model $\theta_K$ and $v_K$; learning rates $\{\eta_t\}_{t=K-B}^{K-1}$;  momentum coefficient $\mu$ ($>0$); minibatches of data $\{D_t\}_{t=K-B}^{K-1}$\;
{\em Initialization}: $\mathrm{d}\theta=\frac{\partial}{\partial \theta_K}\mathcal{M}(\Ddev;\theta_K);\mathrm{d}v=-\eta_K\mathrm{d}\theta_K;\mathrm{d}\omega\leftarrow0;\theta\leftarrow\theta_K;v\leftarrow v_K$\;
\For{$t\leftarrow K-1:-1:K-B$}{
$\theta\leftarrow\theta+\eta_tv$; $g\leftarrow\frac{\partial}{\partial \theta}\big[w_t^\top\ell(D_t;\theta)+\lambda R(\theta)\big]$; $v\leftarrow\frac{v-g}{\mu}$\;
$\mathrm{d}\omega\leftarrow\mathrm{d}\omega+\frac{\partial}{\partial \omega}(g^\top\mathrm{d}v)$; $\mathrm{d}\theta\leftarrow\mathrm{d}\theta+\frac{\partial}{\partial \theta}(g^\top\mathrm{d}v)$; $\mathrm{d}v\leftarrow-\eta_t\mathrm{d}\theta+\mu\mathrm{d}v$\;
}
{\em Return} $\mathrm{d}\omega$.
\caption{The gradients of the validation metric w.r.t. the parameters of the teacher.}
\label{alg:teacher_gradient}
\end{algorithm}

In Algorithm~\ref{alg:teacher_gradient}, we can see that we need a backpropagation interval $B$ as an input, indicating how many internal $\theta_\cdot$'s are used to calculate the gradients of the teacher. When $B=K$, all student models on the optimization trajectory will be leveraged. $B$ balances the tradeoff between efficiency and accuracy. To use Algorithm~\ref{alg:teacher_gradient}, we require $\mu>0$.

As shown in step 2, we first calculate $\frac{\partial}{\partial \theta_K}\mathcal{M}(\Ddev;\theta_K)$, with which we can initialize $\mathrm{d}\theta$, $\mathrm{d}v$ and $\mathrm{d}\omega$. We then recover the $\theta$, $v$ and the gradients at the previous step (see step 4). Based on Eqn.\eqref{eq:derivation}, we recover the corresponding $\mathrm{d}\theta$, $\mathrm{d}v$ and $\mathrm{d}\omega$. We repeat step 4 and step 5 until getting the eventual $\mathrm{d}\omega$, which is the gradient of the validation metric w.r.t. the parameters of the teacher model. Finally, we can leverage any gradient-based algorithm to update the teacher model. With the new teacher model, we can iteratively update $\theta^*$ and $\omega$ until reaching the stopping criteria.

In order to avoid calculating Hessian matrix, which neess to store $O(\vert\theta\vert^2)$ parameters, we leverage the property that $\frac{\partial^2\ell}{\partial \theta^2}v=\frac{\partial}{\partial \theta}g^\top v$, where $\ell$ is the loss function related to $\theta$, $v$ is a vector with size $\vert\theta\vert\times1$, and $g=\frac{\partial \ell}{\partial\theta}$. With this trick, we only require $O(\vert\theta\vert)$ GPU memory.

\noindent{\bf Discussions}: Compared to previous work~\cite{jiang2018mentornet,metaweightnet,fan2018dataTeach,wu2018lossTeach}, except for the key differences that we use internal states as features, there are some differences in optimization. In \cite{fan2018dataTeach}, the teacher is learned in a reinforcement learning manner, which is relatively hard to optimize. In \cite{wu2018lossTeach}, the student model is optimized with vanilla SGD, by which all the intermediate $\theta_t$ should be stored. In our algorithm, we use momentum SGD, where we only need to store the final $\theta_K$ and $v_K$, by which we can recover all intermediate parameters. We will study how to effectively apply our derivations to more optimizers and more applications in the future.

%\noindent(2) In \cite{metaweightnet,ren2018learning,jiang2018mentornet}, once the student model is forwarded once, the teacher model is updated immediately. A potential drawback of such a method is the unstability of the optimization. The optimization of a student model benefits from a stable student which can provide more accurate gradient. Therefore, in our algorithm, we choose to set a larger $K$.}

\subsection{Teacher model}
We introduce the default network architecture of the teacher model used in experiments. We use a linear model with sigmoid activation. Given a pair $(x,y)$, we first use $\varphi_f$ to extract the output of the second-to-last layer, i.e., $I = \varphi_f(x)$. The surface feature $M$ we choose is the one-hot representation of the label, i.e., $M=y$. Then weight of the data $(x,y)$ is  $\varphi(I,M)=\sigma(W_II + EM + b)$, where $\sigma(\cdot)$ denotes the sigmoid function, $W_I$, $E$, and $b$ are the parameters to be learned. $E$ can be regarded as an embedding matrix, which enriches the representations of the labels. One can easily extend the teacher model to a multi-layer feed-forward network by replacing $\sigma$ with a deeper network. 

We need to normalize the weights within a minibatch. When a minibatch $D_t$ comes, after calculating the weight $w_{t,k}$ for the data $(x_{t,k},y_{t,k})\in D_t$, it is normalized as $\tilde{w}_{t,k}=w_{t,k}/\sum_{j=1}^{\vert D_t\vert}w_{t,j}$. This is to ensure that the sum of weights within a batch $D_t$ is always $1$.

\section{Experiments on Image Classification}\label{sec:exp_cifar_clean}
%In this section, we conduct experiments to verify the effectiveness of our proposed method, and we conduct comprehensive ablation studies to explore the importance of components in the method.

In this section, we conduct experiments on CIFAR-$10$ and CIFAR-$100$ image classification. We first show the overall results and then provide several analysis. Finally, we apply our algorithm to the image classification with noised labels.

\subsection{Settings}

There are $50000$ and $10000$ images in the training and test sets. CIFAR-$10$ and CIFAR-$100$ are a $10$-class and a $100$-class classification tasks respectively. We split $5000$ samples from the training dataset as $\Ddev$ and the remaining $45000$ samples are used as $\Dtrain$. Following~\cite{he2016Resnet}, we use momentum SGD with learning rate $0.1$ and divide the learning rate by $10$ at the $80$-th and $120$-th epoch.
The momentum coefficient $\mu$ is $0.9$. The $K$ and $B$ in Algorithm~\ref{alg:teacher_gradient} are set as $20$ and $2$ respectively. We train the models for $300$ epochs to ensure convergence. The minibatch size is $128$. We conducted experiments on ResNet-32, ResNet-110 and Wide ResNet-28-10 (WRN-28-10)~\cite{BMVC2016_87}. All the models are trained on a single P40 GPU.

We compare the results with the following baselines: (1) The baseline of data teaching~\cite{fan2018dataTeach} and loss function teaching~\cite{wu2018lossTeach}. They are denoted as L2T-data and L2T-loss respectively. (2) Focal loss~\cite{lin2017focal}, where each data is weighted by $(1-p)^\gamma$, $p$ is the probability that the data is correctly classified, and $\gamma$ is a hyperparameter. We search $\gamma$ on $\{0.5,1,2\}$ suggested by~\cite{lin2017focal}. (3) Self-paced learning (SPL)~\cite{Kumar2010SPL}, where we start from easy samples first and then move to harder examples. 

% For the teacher model, we use a linear model with sigmoid activation. That is, $\varphi(I,M)=\sigma(W_II + W_MEM + b)$, where $\sigma(\cdot)$ denotes the sigmoid function, $W_I$, $W_M$, $E$, and $b$ are the parameters to be learned. For $I$, we use the output from the second-to-last layer of the student models, whose dimensions are $64$, $64$ and $640$ for ResNet-32, ResNet-110 and WRN-28-10. $M$ is the one-hot representation of the label, and $E$ is the corresponding learnable embedding matrix. 

\subsection{Results}

The test error rates of different settings are reported in Table~\ref{tab:CIFAR-10-100-nomral-setting}. For CIFAR-$10$, we can see that the baseline results of ResNet-32, ResNet-110 and WRN-28-10 are $7.22$, $6.38$ and $4.27$ respectively. With our method, we can obtain $6.20$, $5.65$ and $3.72$ test error rates, which are the best  among all listed algorithms. For CIFAR-100, our approach can improve the baseline by $0.92$, $1.67$ and $1.11$ points. These consistent improvements demonstrate the effectiveness of our method. We have the following observations: (1) L2T-data is proposed to speed up the training. Therefore, we can see that the error rates are almost the same as the baselines. (2) For L2T-loss, on CIFAR-10 and CIFAR-100, it can achieve $0.27$ and $0.32$ points improvements, which are far behind of our proposed method. This shows the great advantage of our method than the previous learning to teach algorithms. (3) Focal loss sets weights to the data according to the hardness only, which does not leverage internal states neither. There exists non-negligible gap between focal loss and our method. (4) For SPL, the results are similar (even worse) to the baseline. This shows the importance of a learning based scheme for data selection.

\begin{table*}[!htbp]
\centering
\begin{tabular}{l|cccccccc}
\toprule
CIFAR-$10$ & Baseline & L2T-data & L2T-loss & Focal loss & SPL & Ours  \\
\midrule
ResNet-32    & $7.22$ & $7.16$ & $6.95$ & $6.60$ & $11.48$ & $6.20$ \\
ResNet-110   & $6.38$ & $6.10$ & $6.02$ & $6.19$ & $11.06$ & $5.65$ \\
WRN-28-10    & $4.27$ & $4.09$ & $3.97$ & $4.57$ & $4.25$ & $3.72$ \\
\midrule
CIFAR-$100$ & Baseline & L2T-data & L2T-loss & Focal loss & SPL & Ours  \\
\midrule
ResNet-32    & $29.57$ & $29.54$ & $29.25$ & $28.85$ & $29.98$ & $28.65$ \\
ResNet-110   & $27.69$ & $27.02$ & $26.61$ & $26.55$ & $27.91$ & $26.02$\\
WRN-28-10    & $20.49$ & $19.92$ & $19.93$ & $19.86$ & $20.56$ & $19.38$ \\
\bottomrule
\end{tabular}
\caption{Results on CIFAR-$10$/CIFAR-$100$. The labels are clean.}
\label{tab:CIFAR-10-100-nomral-setting}
\end{table*}

\subsection{Analysis}
\label{sec:analysis}
To further verify how our method works, we conduct several ablation studies. All experiments are conducted on CIFAR-$10$ with ResNet-32.

\noindent{\em Comparison with surface information}: The features of the teacher model used in Table~\ref{tab:CIFAR-10-100-nomral-setting} are the output of the second-to-last layer of the network (denoted as $I_0$), and the label embedding (denoted as $M_0$). Based on~\cite{metaweightnet,ren2018learning,wu2018lossTeach,fan2018dataTeach}, we define another group of features about surface information.  Five components are included: the training iteration (normalized by the total number of iteration), average training loss until the current iteration, best validation accuracy until the current iteration, the predicted label of the current input, and the margin values. These surface features are denoted as $M_1$.

For the teacher model, We try different combinations of the internal states and surface features. The settings and results are shown in Table~\ref{tab:ablation-meta-feature}.

\begin{table}[!htbp]
\centering
\begin{tabular}{l|c}
\toprule
Setting & Error rate \\
\midrule
$I_0 + M_0$ & $6.20$ \\
$I_0$ & $6.34$ \\
$M_0$ & $6.50$ \\
$M_1$ & $6.54$ \\
$M_0 + M_1$ & $6.50$ \\
$I_0 + M_0 + M_1$ & $6.30$ \\
\bottomrule
\end{tabular}
\caption{Ablation study on the usage of features.}
\label{tab:ablation-meta-feature}
\end{table}

As shown in Table~\ref{tab:ablation-meta-feature}, we can see that the results of using surface features only (i.e., the settings without $I_0$) cannot catch up with those with internal states of the network (i.e., the settings with $I_0$). This shows the effectiveness of the internal states for learning to teach. We do not observe significant differences among the settings $M_0$, $M_1$ and $M_0+M_1$. 
%Therefore, we choose the use the easiest one, $M_0$, which requires less computation cost than $M_1$.
Using $I_0$ only can result in less improvement than using $I_0 + M_0$. Combining $I_0$, $M_0$ and $M_1$ also slightly hurts the result. Therefore, we choose $I_0 + M_0$ as the default setting.

\noindent{\em Internal states from different levels}: By default, we use the output of second-to-last layer as the features of internal states. We also try several other variants, naming $I_1$, $I_2$ and $I_3$, which are the outputs of the last convolutional layer with size $8\times8$, $16\times16$ and $32\times32$. A larger subscript represents that the corresponding features are more similar to the raw input. We explore the setting $I_i + M_0$, $i\in\{0,1,2,3\}$. Results are reported in Table~\ref{tab:ablation_feature_diff_level}. We can see that leveraging internal states (i.e., $I_\cdot$) can achieve lower test error rates than those without such features. Currently, there is not significant difference on where the internal states are from. Therefore, by default, we recommend to use the states from the second-to-last layer. 

\begin{table}[!htbp]
\centering
\begin{tabular}{ccccc}
\toprule
Setting ($I_i + M_0$)    &  $0$ & $1$ & $2$ & $3$ \\
\midrule
Error rate & $6.20$  & $6.22$ & $6.31$ & $6.21$\\
\bottomrule
\end{tabular}
\caption{Features from different levels.}
\label{tab:ablation_feature_diff_level}
\end{table}
\begin{table}
\centering
\begin{tabular}{cccc}
\toprule
Setting & MLP-0 &  MLP-1  & MLP-2  \\
\midrule
Error rate & $6.20$ &  $6.48$ &  $6.59$   \\
\bottomrule
\end{tabular}
\caption{Teacher with various hidden layers.}
\label{tab:ablation_teacher_net}
\end{table}

\noindent{\em Architectures of the teacher models}: We explore the teacher networks with different number of hidden layers. Each hidden layer is followed by a ReLU activation (denoted as MLP-\#layer). The dimension of the hidden states are the same as the input. Results are in Table~\ref{tab:ablation_teacher_net}.

Using a more complex teacher model will not bring improvement to the simplest one as we used in the default setting. Our conjecture is that more complex models are harder to optimize, which can not provide accurate signals for the student models. 

\begin{figure*}[!htbp]
\centering
\begin{minipage}{0.31\linewidth}
\subfigure[Weight-loss curve, $\mathcal{T}_0$]{
\includegraphics[width=\linewidth]{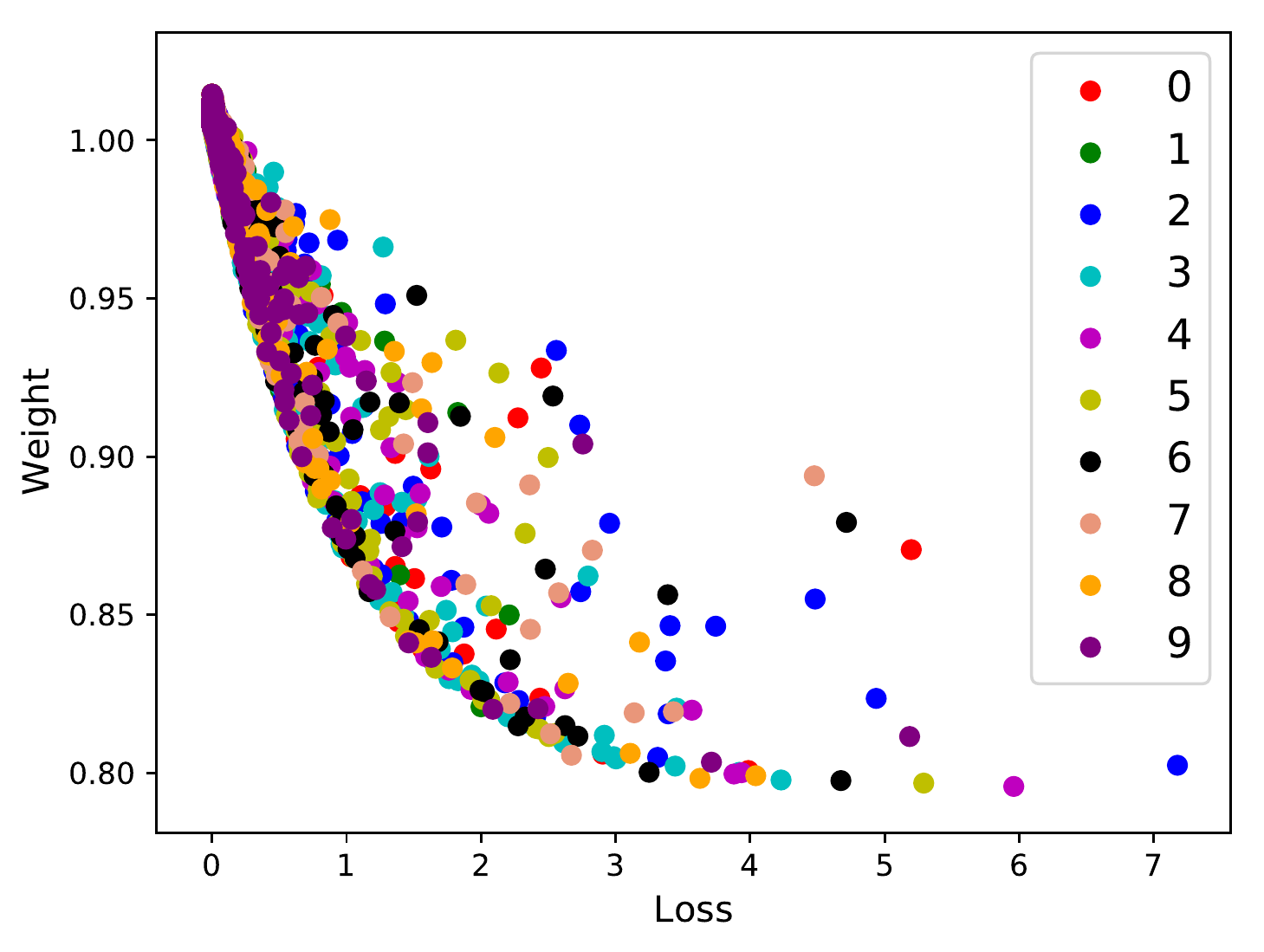}
}
\end{minipage}%
\begin{minipage}{0.31\linewidth}
\subfigure[Internal features, $\mathcal{T}_0$]{
\includegraphics[width=\linewidth]{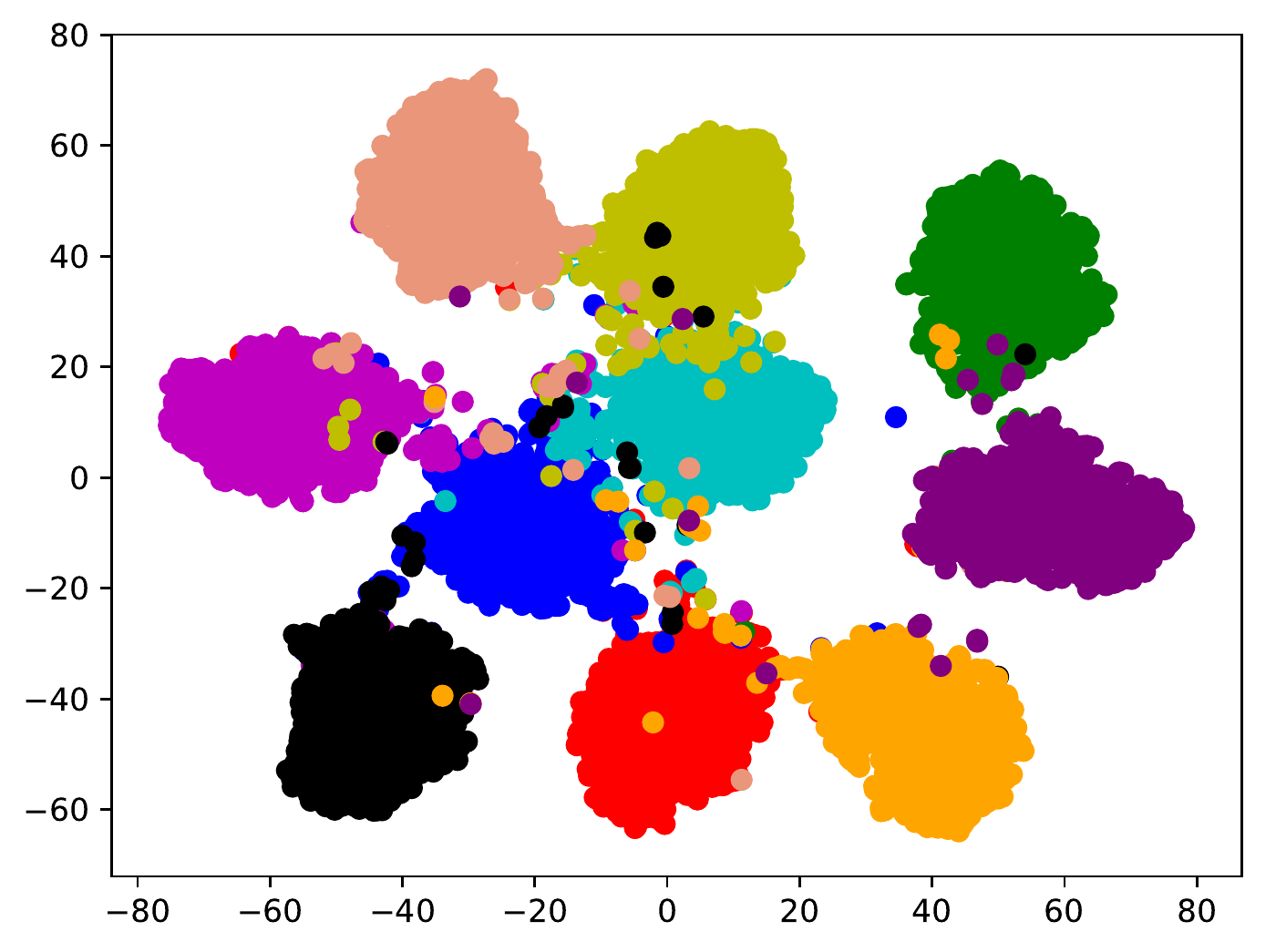}
}
\end{minipage}%
\begin{minipage}{0.31\linewidth}
\subfigure[Weight w.r.t. classes, $\mathcal{T}_0$]{
\includegraphics[width=\linewidth]{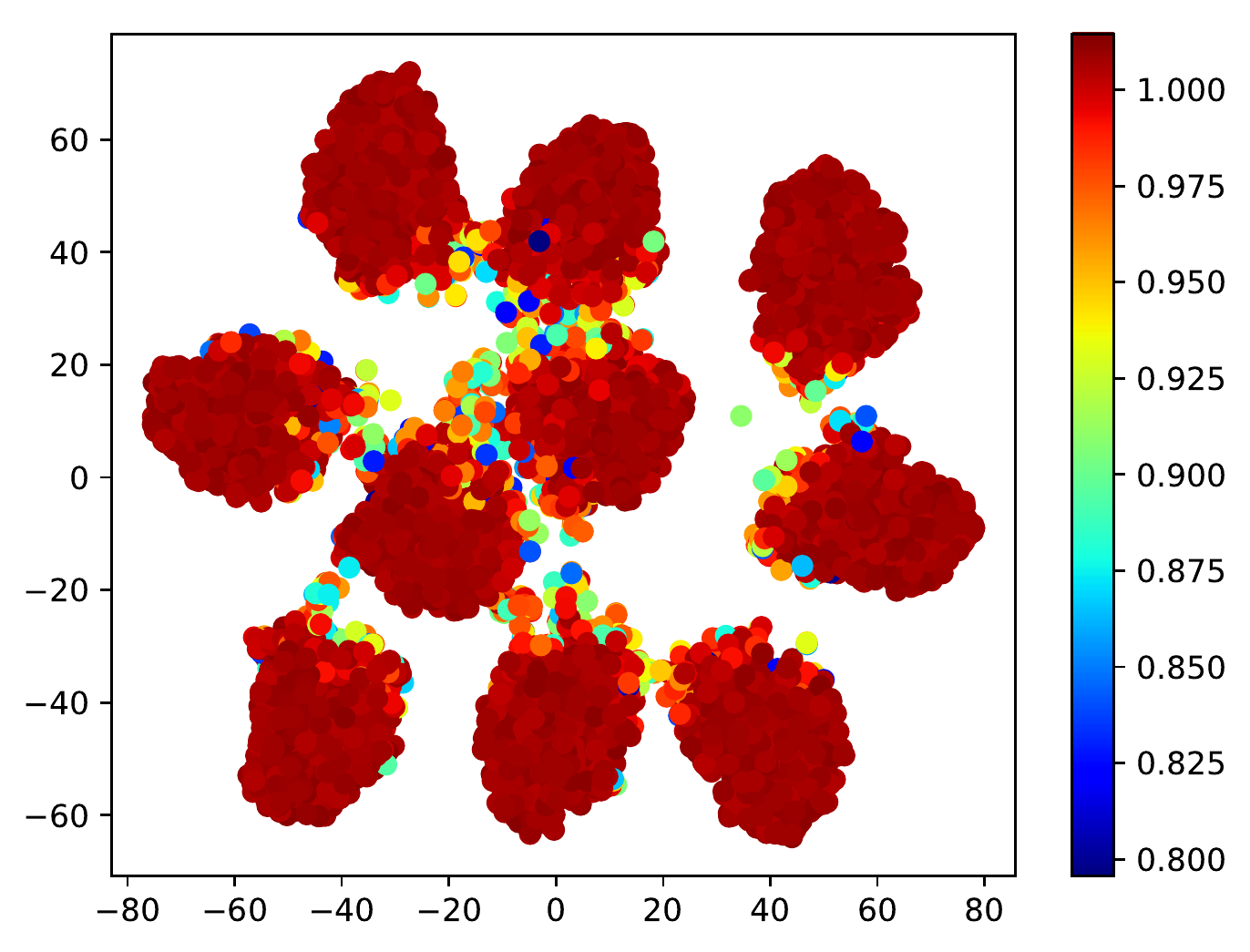}
}
\end{minipage}
\begin{minipage}{0.31\linewidth}
\subfigure[Weight-loss curve, $\mathcal{T}_1$]{
\includegraphics[width=\linewidth]{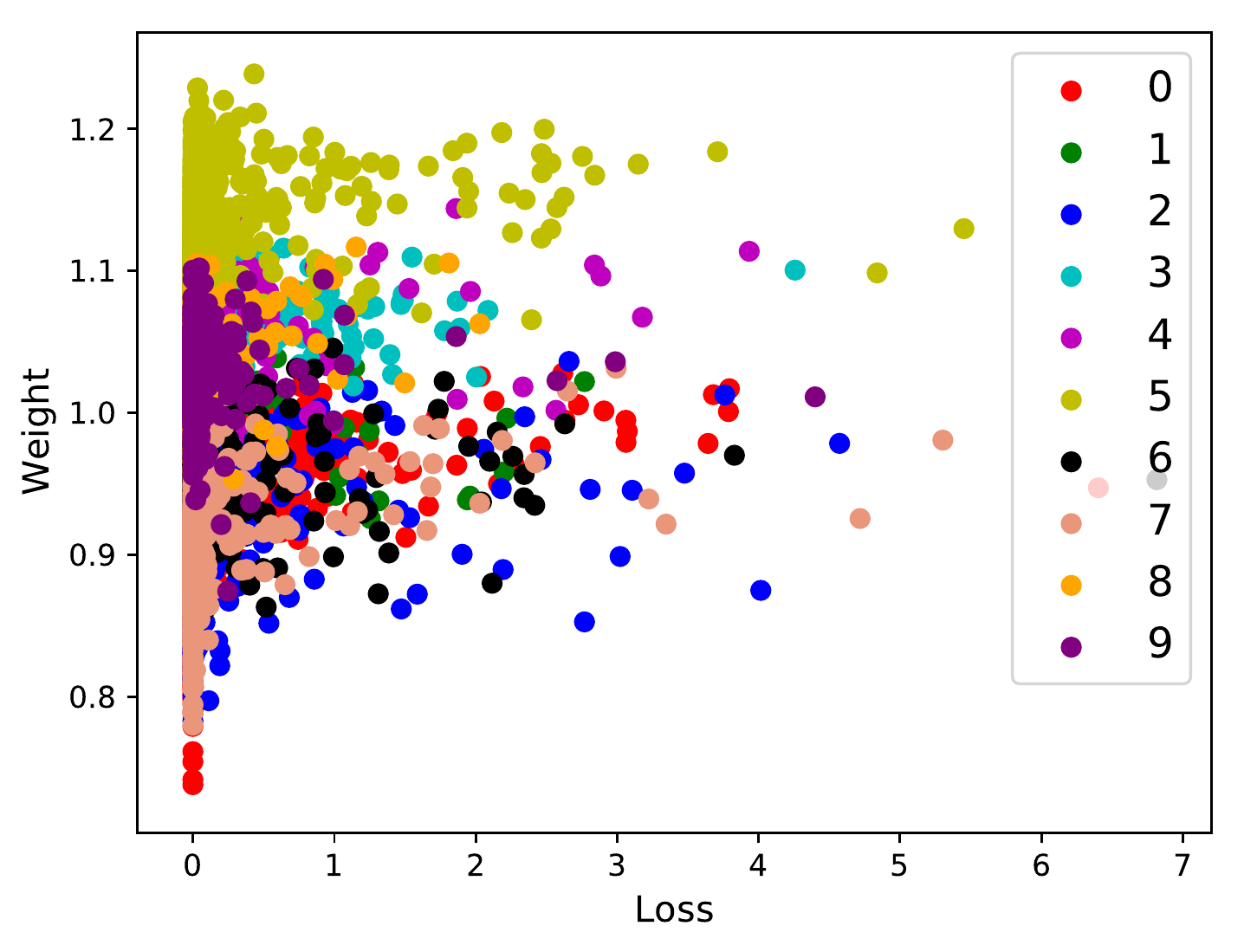}
}
\end{minipage}%
\begin{minipage}{0.31\linewidth}
\subfigure[Internal features, $\mathcal{T}_1$]{
\includegraphics[width=\linewidth]{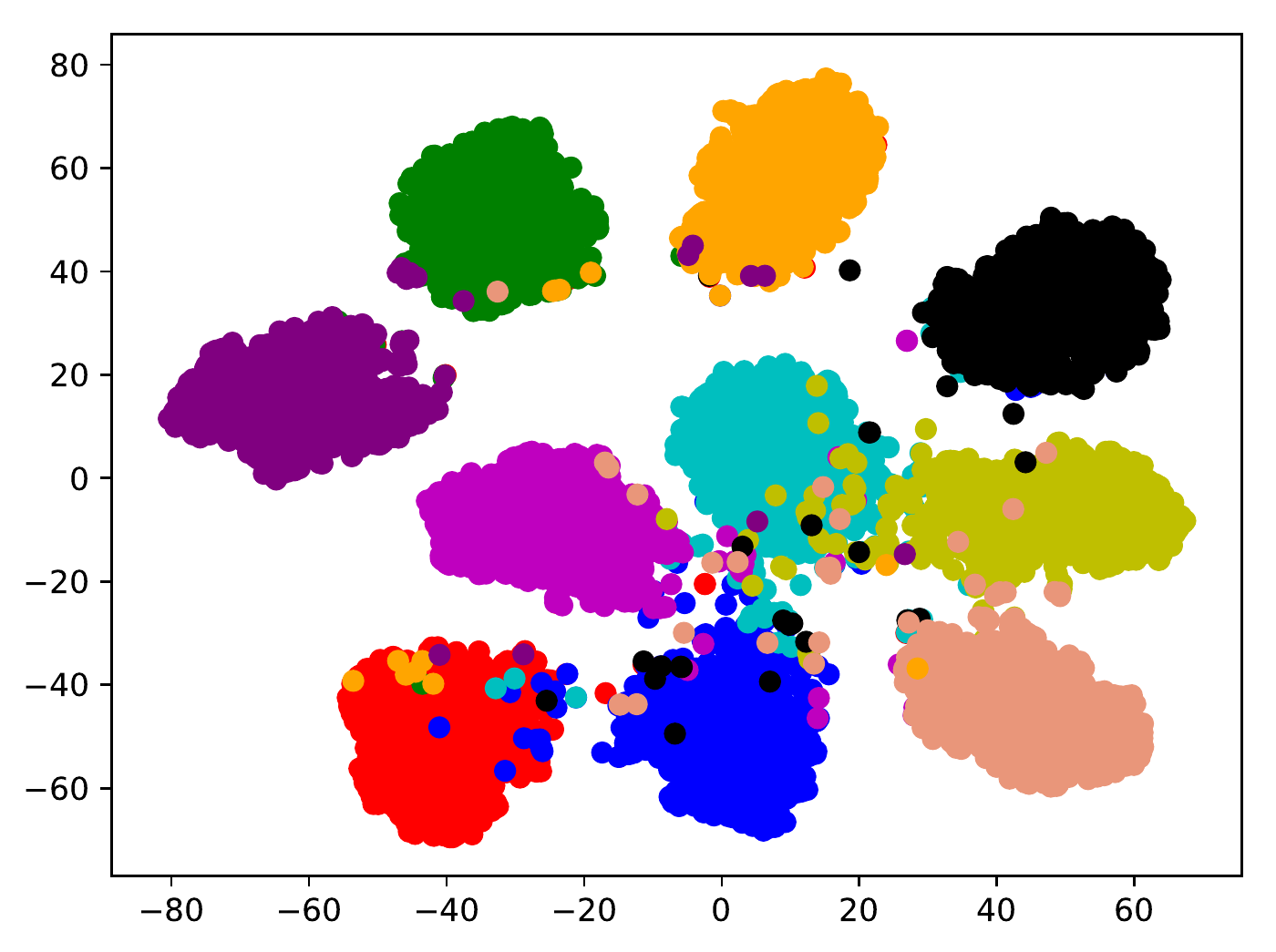}
}
\end{minipage}%
\begin{minipage}{0.31\linewidth}
\subfigure[Weight w.r.t. classes, $\mathcal{T}_1$]{
\includegraphics[width=\linewidth]{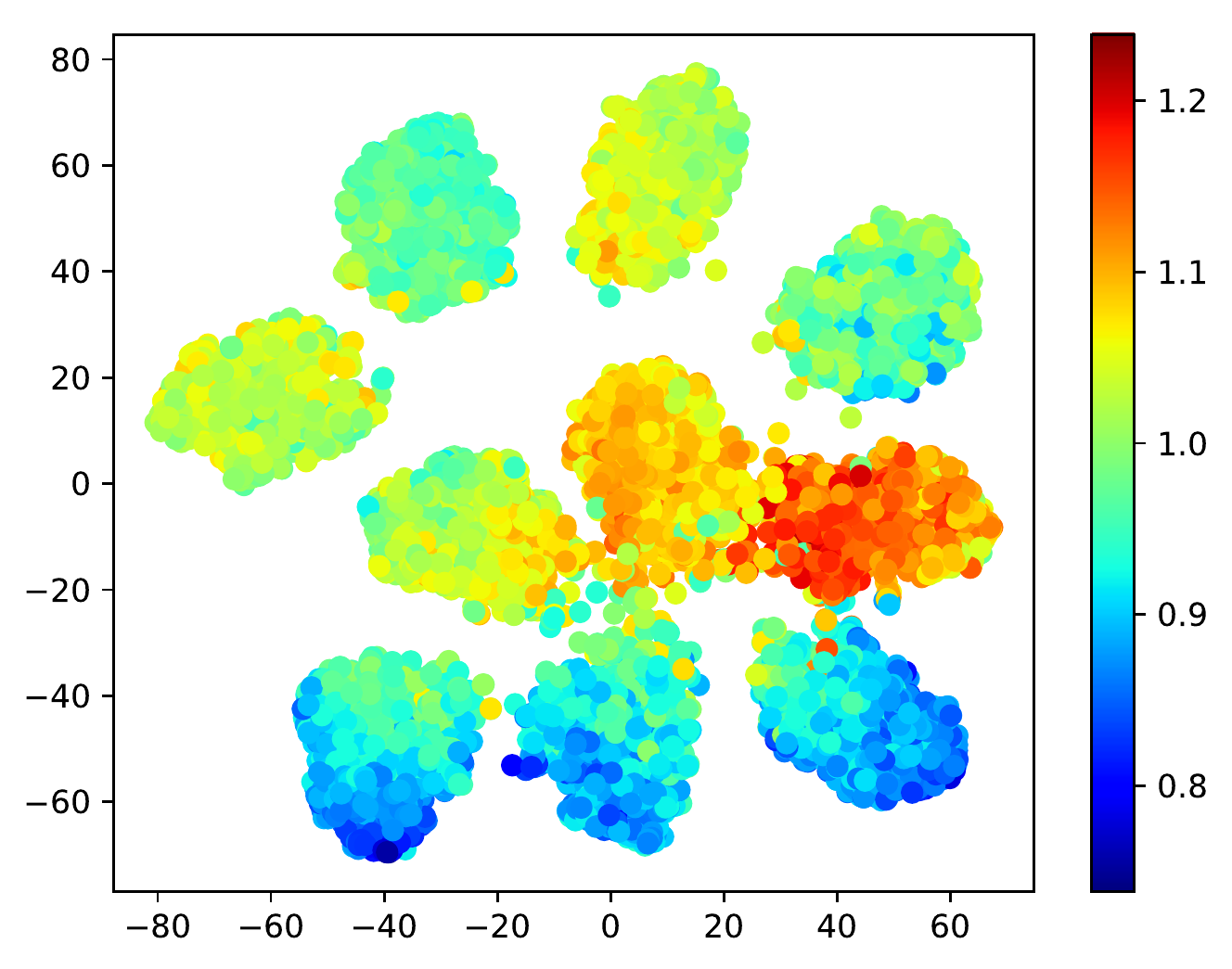}
}
\end{minipage}
\caption{Visualization of weights and loss values of $\mathcal{T}_0$ and $\mathcal{T}_1$.
%In (a), (b), (d) and (e), the data points in the same category are painted with the same color.
%(a) and (d) show the correlation between the output data weight and the training loss;
%(b) and (e) visualize the internal features of all data points;
%(c) and (f) show heatmaps regarding output weights of all data points.
}
\label{fig:viz_weights}
\end{figure*}

\noindent{\em Analysis on the weights}: We take comparison between the weights output by the teacher model leveraging surface features $M_1$ only (denoted as $\mathcal{T}_0$) and those output by our teacher leveraging internal features (denoted as $\mathcal{T}_1$). The results are shown in Figure~\ref{fig:viz_weights}, where the top row represents the results of $\mathcal{T}_0$ and the bottom row for $\mathcal{T}_1$. In Figure~\ref{fig:viz_weights}(a), (b), (d), (e), the data points in the same category are painted with the same color. The first column shows the correlation between the output data weight ($y$-axis) and the training loss ($x$-axis); the second column is used to visualize the internal states through t-SNE~\cite{maaten2008visualizing}; the third column plots heatmaps regarding output weights of all data points (red means large weight and blue means smaller), in accordance with those in the second column.  We have the following observations: 

\noindent(1) As shown in the first column, $\mathcal{T}_0$ tries to assign lower weights to the data with higher loss, regardless of the category the image belongs to. In contrast, the weights set by $\mathcal{T}_1$ heavily rely on the category information. For example, the data points with label $5$ have the highest weights regardless of the training loss, followed by those with label $3$, where label $3$ and $5$ correspond to the ``cat'' and ``dog'' in CIFAR-10, respectively. 

\noindent(2) To further investigate the reason why the data of cat class and dog class are assigned with larger weights by $\mathcal{T}_1$, we turn to Figure~\ref{fig:viz_weights}(e), from which we can find that the internal states of dog and cat are much overlapped. We therefore hypothesize that, since the dog and cat are somewhat similar to each other, $\mathcal{T}_1$ is learned to separate these two classes by assigning large weights to them. Yet, this phenomenon cannot be observed in $\mathcal{T}_0$.

% \noindent(3) Finally, as shown in Figure~\ref{fig:viz_weights}(b) and (e), we observe many abnormal points wrongly lying in the cluster of other classes. The loss values of abnormal points are generally larger compared to the normal ones. With $\mathcal{T}_0$, they are assigned as relatively low weights, which means to pay less attention to them. In contrast, with $\mathcal{T}_1$, they obtain larger weights, indicating that the teacher wants to ``pull'' them back to the correct cluster.

% , followed by the third class (i.e., cat). In the third class, even if for the data with high loss values, it can still be assigned larger weights. For the points which are misclassified into other classes, $\mathcal{T}_0$ assigns low weights to them since they have large losses. In contrast, these points have relatively larger weight with $\mathcal{T}_1$.

% \noindent(2) In Figure~\ref{fig:viz_weights}(e) and (f), we know that weights of third (cat) and fifth (dog) classes are the highest. The images of these two classes are similar. Therefore, $\mathcal{T}_1$ assigns larger weights to make them distinguishable. 

\noindent{\it Preliminary exploration on deeper interactions}: To stabilize training, we do not backpropagate the gradients to the student model via the weights, i.e., $\frac{\partial w_t}{\partial \theta_t}$ is set as zero.  If we enable $\frac{\partial w_t}{\partial \theta_t}$, the teacher model will have another path to pass the supervision signal to the student model, which has great potential to improvement the performances. We quickly verify this variant on CIFAR-$10$ using ResNet-32. We choose $I_0 + M_0$ as the features of the teacher model.%, and choose MLP-1 + Attn as the teacher model architecture. 
We find that with this technique, we can further lower the test error rate to $6.08\%$, another $0.12$ improvement compared to the current methods. We will further exploration this direction in the future.

\subsection{Image Classification with Noisy Labels}\label{sec:exp_cifar_noise}

To verify the ability of our proposed method to deal with the noisy data, we conduct several experiments on CIFAR-10/100 datasets with noisy labels.

We derive most of the settings from~\cite{metaweightnet}. The images remain the same as those in standard CIFAR-10/100, but we introduce noise to their labels, including the uniform noise and flip noise. For the validation and test sets, both the images and the labels are clean.

\begin{enumerate}
\item {\it Uniform noise}: We follow a common setting from~\cite{zhang2017understanding}. The label of each image is uniformly mapped to a random class with probability $p$. In our experiments, we set the probability $p$ as $40\%$ and $60\%$. Following~\cite{metaweightnet}, the network architecture of the student network is WRN-28-10. We use momentum SGD with learning rate $0.1$ and divide the learning rate by $10$ at $36$-th epoch and $38$-th epoch ($40$ epoch in total).
\item {\it Flip noise}: We follow~\cite{metaweightnet} to set flip noise. The label of each image is independently flipped to two similar classes with probability $p$. The two similar classes are randomly chosen, and we flip labels to them with equal probability. In our experiments, we set probability $p$ as $20\%$ and $40\%$ and adopt ResNet-32 as the student model. We use momentum SGD with learning rate $0.1$ and divide the learning rate by $10$ at $40$-th epoch and $50$-th epoch ($60$ epoch in total).
\end{enumerate}

For the teacher model, we follow settings as those used for clean data. We compare the results with MentorNet~\cite{jiang2018mentornet} and Meta-Weight-Net~\cite{metaweightnet}.

The results are shown in Table~\ref{tab:CIFAR-10-100-uniform-noise-label} and Table~\ref{tab:CIFAR-10-100-flip-noise-label}. Our results are better than previous baselines like MentorNet and Meta-Weight-Net, regardless of the type and magnitude. When the noise type is uniform, we can improve Meta-Weight-Net by about $0.5$ point. On flip noise with ResNet-32 network, the improvement is more significant, where in most cases, we can improve the baseline by more than one point.  The experiment results demonstrate that leveraging internal states is also useful for the datasets with noisy labels. This shows the generality of our proposed method.

\begin{table}[!t]
	\small
	\centering
	\begin{tabular}{l|cc}
		\toprule
		& \multicolumn{2}{c}{CIFAR-$10$} \\
		Method & $p=40\%$ & $p=60\%$ \\
		\midrule
		Baseline & $31.93$ & $46.88$ \\
		MentorNet~\cite{jiang2018mentornet} & $12.67$ & $17.20$ \\
		Meta-Weight-Net~\cite{metaweightnet} & $10.73$ & $15.93$ \\
		Ours & $10.29$ & $15.37$ \\
		\midrule
		& \multicolumn{2}{c}{CIFAR-$100$} \\
		Method & $p=40\%$ & $p=60\%$ \\
		\midrule
		Baseline &  $48.89$ & $69.08$ \\
		MentorNet & $38.61$ & $63.13$ \\
		Meta-Weight-Net & $32.27$ & $41.25$ \\
		Ours & $31.36$ & $40.62$ \\
		\bottomrule
	\end{tabular}
	\caption{Results of WRN-28-10 with uniform noise labels.}
	\label{tab:CIFAR-10-100-uniform-noise-label}
\end{table}

\begin{table}[!h]
	\small
	\centering
	\begin{tabular}{l|cc}
		\toprule
		& \multicolumn{2}{c}{CIFAR-$10$} \\
		Method & $p=20\%$ & $p=40\%$ \\
		\midrule
		Baseline & $23.17$ & $29.23$ \\
		MentorNet~\cite{jiang2018mentornet} & $13.64$ & $18.24$ \\
		Meta-Weight-Net~\cite{metaweightnet} & $9.67$ & $12.46$ \\
		Ours & $8.95$ & $11.29$ \\
		\midrule
		& \multicolumn{2}{c}{CIFAR-$100$} \\
		Method & $p=20\%$ & $p=40\%$ \\
		\midrule
		Baseline & $49.14$ & $56.99$ \\
		MentorNet & $38.03$ & $47.34$ \\
		Meta-Weight-Net & $35.78$ & $41.36$ \\
		Ours & $33.92$ & $39.49$ \\
		\bottomrule
	\end{tabular}
	\caption{Results of ResNet-32 with flip noise labels.}
	\label{tab:CIFAR-10-100-flip-noise-label}
\end{table}

\section{Experiments on  Machine Translation}

In this section, we verify our algorithm on neural machine translation (NMT). We conduct experiments on IWSLT'14 German-to-English (briefly, De$\to$En) translation, with both cleaned and noisy data.

\subsection{Settings}
%\noindent{\em Data preprocessing}:
There are $153k$/$7k$/$7k$ sentence pairs in the training/valid/test sets of IWSLT'14 De$\to$En translation dataset. We first tokenize the words and then leverage byte-pair-encoding (BPE)~\cite{BPE} to split words into sub-word units on the dataset. We use a joint dictionary for the two languages and the vocabulary size is $10k$.

To create the noisy IWSLT'14 De$\to$En dataset, we add noise to each sentence pair with probability $p$. If a sentence is selected to add noise, each word in the source language sentence and the target language sentence is uniformly replaced with a special token ``<MASK>'' with probability $q$. In our experiment, we set $p$ and $q$ as $0.1$ and $0.15$ respectively.

% \noindent{\em Model configuration and hyperparameters}:
For all translation tasks, we reuse the settings in image classification to train the teacher model, and use Transformer~\cite{Transformer} as the student model. We derive most settings from the fairseq official implementation\footnote{\url{https://github.com/pytorch/fairseq/blob/master/fairseq/models/transformer.py}} and use the \texttt{ transformer\_small} configuration for the student model, where the embedding dimension, hidden dimension of feed-forward layers and number of layers are $512$, $1024$ and $6$ respectively. We first use Adam algorithm with learning $5\times10^{-4}$ to train an NMT model until convergence, which takes about one day. Then we use pre-trained models to initialize the student model and use momentum SGD for finetuning with learning rate $10^{-3}$, which takes about three hours. The batchsize is $4096$ tokens per GPU.  We implement the data teaching~\cite{fan2018dataTeach} as a baseline, which is denoted as L2T-data.
We use BLEU score~\cite{BLEU} as the evaluation metric, which is calculated by  \texttt{multi-bleu.perl}\footnote{\url{https://github.com/moses-smt/mosesdecoder/blob/master/scripts/generic/multi-bleu.perl}}. 

\subsection{Results}

The BLEU score results of neural machine translation tasks are reported in Table~\ref{tab:nmt_exp}. We can see that our proposed method can achieve more than $1$ points gain on all translation tasks compared with the baseline, and also outperforms the previous approach of L2T-data. For noisy IWSLT'14 De$\to$En task, our approach can improve the baseline by $1.88$ points, which indicates that our proposed method is more competitive on noisy datasets.

\begin{table}[!htbp]
    \centering
    \begin{tabular}{@{}l|p{0.99cm}p{1.35cm}p{.68cm}}
        \toprule
        Task & Baseline & L2T-data & Ours \\
        \midrule
        IWSLT'14 De$\to$En (clean) & $34.95$ & $35.61$ & $36.00$ \\
        IWSLT'14 De$\to$En (noisy) & $33.68$ & $34.42$ & $35.56$ \\
        % WMT'14 En$\to$De & & & \\
        \bottomrule
    \end{tabular}
    \caption{BLEU scores on NMT tasks.}
    \label{tab:nmt_exp}
\end{table}

\section{Conclusion and Future Work}\label{sec:conclusion}
We propose a new data teaching paradigm, where the teacher and student model have deep interactions. The internal states are fed into the teacher model to calculate the weights of the data, and we propose an algorithm to jointly optimize the two models. Experiments on CIFAR-$10$/$100$ and neural machine translation tasks with clean and noisy labels demonstrate the effectiveness of our approach. Rich ablation studies are conducted in this work. 

For future work, the first is to study how to apply deeper interaction to the learning to teach framework (preliminary results in Section~\ref{sec:analysis}). Second, we want that the teacher model could be transferred across different tasks, which is lacked for the current teacher (see Appendix B for the exploration). Third, we will carry out theoretical analysis on the convergence of the optimization algorithm.

{\fontsize{9.0pt}{10.0pt} 
\selectfont 
\bibliography{mybib}
}

\appendix
\section{Ablation Study on Different $K$ and $B$}

In this section, we conduct an ablation study to explore the impact of different model update interval $K$ and backpropagation interval $B$ on our algorithm. We adopt CIFAR-$10$ and ResNet-32 as the base dataset and student model respectively. We choose $(K, B) \in \{(1, 1), (20, 2), (20, 5), (100, 2), (100, 5)\}$ in our ablation study. In Table~1 in the main paper, we use $K=20$ and $B=2$ as the default setting.

The ablation study results are reported in Table~\ref{tab:ablation-K-B}. We can observe that 1) The setting that run backpropagation at each step ($K=1, B=1$) takes high computational cost and is hard to optimize the teacher model. 2) Our default setting can reach the lowest test error rate among all $(K, B)$ settings.

\begin{table}[!htbp]
	\centering
	\begin{tabular}{ll|c}
		\toprule
		$K$ & $B$ & Test error  \\
		\midrule
		$1$ & $1$ & $6.56$ \\
		$20$ & $2$ & $6.20$ \\
		$20$ & $5$ & $6.41$ \\
		$100$ & $2$ & $6.30$ \\
		$100$ & $5$ & $6.24$ \\
		\bottomrule
	\end{tabular}
	\caption{Error rates of different $K$ and $B$ on CIFAR-$10$ and student model ResNet-32.}
	\label{tab:ablation-K-B}
\end{table}

\section{Transferability of the Teacher across Different Tasks}

In this section, we conduct some experiments to explore the transferability of our teacher models across different tasks.

We choose our best setting $6.20$ (the dataset is CIFAR-$10$, and the architecture of the student model is ResNet-32) in Table~1 as the original teacher model, and adopt two transfer settings.

\noindent(1) {\em{Transfer to different dataset}}: We transfer our original teacher model from CIFAR-$10$ to CIFAR-$100$ dataset. The network architecture of the student model remains unchanged.

\noindent(2) {\em{Transfer to different student model}}: We change the student model architecture from ResNet-32 to ResNet-110. The dataset remains unchanged.

In the above two settings, we train the student models from scratch and fix the parameters of the teacher models. The teacher models provide weights for the input data.
%After transferring, we retrain the student model from scratch, and use the transferred teacher to produce the weight of each data sample. 
%In this retraining procedure, we disable the backpropagation steps, so the teachers are ``frozen'' and will not be updated.

The test error rates of \textit{Transfer to different dataset} and \textit{Transfer to different student model} are $92.45\%$ and $57.97\%$ respectively. Our teacher models lack transferability due to deep interactions between the teacher and the student models. We will improve our algorithm to enhance the transferability in future.

\end{document}